\documentclass[conference]{IEEEtran}
\IEEEoverridecommandlockouts

\usepackage{cite}
\usepackage{float}
\usepackage{amsmath,amssymb,amsfonts}
\usepackage{algorithmic}
\usepackage{graphicx}
\usepackage{textcomp}
\usepackage{xcolor}
\usepackage{hyperref}
\usepackage{bbm}
\usepackage{booktabs}
\usepackage{tikz}
\usetikzlibrary{arrows.meta,positioning,fit}

\def\BibTeX{{\rm B\kern-.05em{\sc i\kern-.025em b}\kern-.08em
    T\kern-.1667em\lower.7ex\hbox{E}\kern-.125emX}}

\newcommand{\pipeImgHeight}{1.9cm}

\begin{document}

\title{MineXplore: An Open-Source Reinforcement Learning Exploration
Benchmark for GNSS-Denied Underground Environment}

\author{
\IEEEauthorblockN{Abhishek Seetharamu\textsuperscript{*}}
\IEEEauthorblockA{\textit{BuildMachineLabs}\\
abhishekss6363@gmail.com}
\and
\IEEEauthorblockN{Badrikanath Praharaj\textsuperscript{*}}
\IEEEauthorblockA{\textit{BuildMachineLabs}\\
badrikanath.praharaj@gmail.com}
\and
\IEEEauthorblockN{Sreeram M.V.\textsuperscript{*}}
\IEEEauthorblockA{\textit{BuildMachineLabs}\\
mvsreeramblr@gmail.com}
\thanks{\textsuperscript{*}These authors contributed equally to this work.}
}
\maketitle

\begin{abstract}
Underground mines present extreme conditions for autonomous robot
navigation: GPS is denied, lighting is degraded, and tunnel topology is
loop-rich and non-convex. Simulation benchmarks grounded in real
production-mine geometry and compatible with GPU-accelerated learning
pipelines do not yet exist in the open-source ecosystem. We present
MineXplore, an open-source MuJoCo-based navigation benchmark derived
from the Leung et al.\ 2017 Chilean underground copper mine dataset
\cite{leung2017}. The environment reconstructs a 104{,}423\,m$^2$ tunnel
network through an six-stage contour-to-MJCF pipeline incorporating
octagonal wall cross-sections, LiDAR-sourced jagged wall geometry, three
terrain friction zones, a global 5$^\circ$ incline, and periodic spot
lighting. Geometric fidelity is validated at an Intersection over Union
(IoU) of 0.9538 against the source survey map, and surface texture
similarity scores 79.4\% across six structural dimensions. A
single-agent PPO baseline trained via RLlib  across five independent random seeds achieves a best
rolling coverage of $88.89\%\ {\pm}\ 1.74\%$ (3 of 5 seeds reaching the
90\% coverage target), confirming that MineXplore supports stable and
reproducible policy learning under realistic underground sensing and
topology.
\end{abstract}

\begin{IEEEkeywords}
simulation environment, underground mining robotics, reinforcement
learning, MuJoCo, navigation benchmark, exploration
\end{IEEEkeywords}

\section{Introduction}

Underground mines are among the hardest environments for autonomous
ground robots. GPS is denied, lighting is degraded, dust and water reduce
sensor returns, and topology is loop-rich and non-convex
\cite{ebadi2024}. The DARPA Subterranean Challenge surfaced both the
difficulty of such settings and the community's best field systems
\cite{tranzatto2022}. Yet much of the simulation tooling used for
algorithm development remains centred on Gazebo, even as GPU-accelerated
simulators have become the default for modern robot learning pipelines
\cite{makoviychuk2021,zakka2025}.

Two gaps motivate this work. First, no publicly released MuJoCo or MJX
environment is grounded in a real production mine; existing subterranean
simulation assets are predominantly Gazebo-based \cite{subt2021,gao2025}.
Second, rare published real-mine datasets such as the Leung et al.\ 2017
Chilean underground mine dataset have been used almost exclusively for
SLAM evaluation rather than as geometric sources for simulation
\cite{leung2017,ebadi2024}. MineXplore closes both gaps by compiling the
Chilean survey data into a calibrated MuJoCo world directly usable for
reinforcement learning under local observations.

\begin{figure}[t]
  \centering
  \includegraphics[width=\columnwidth]{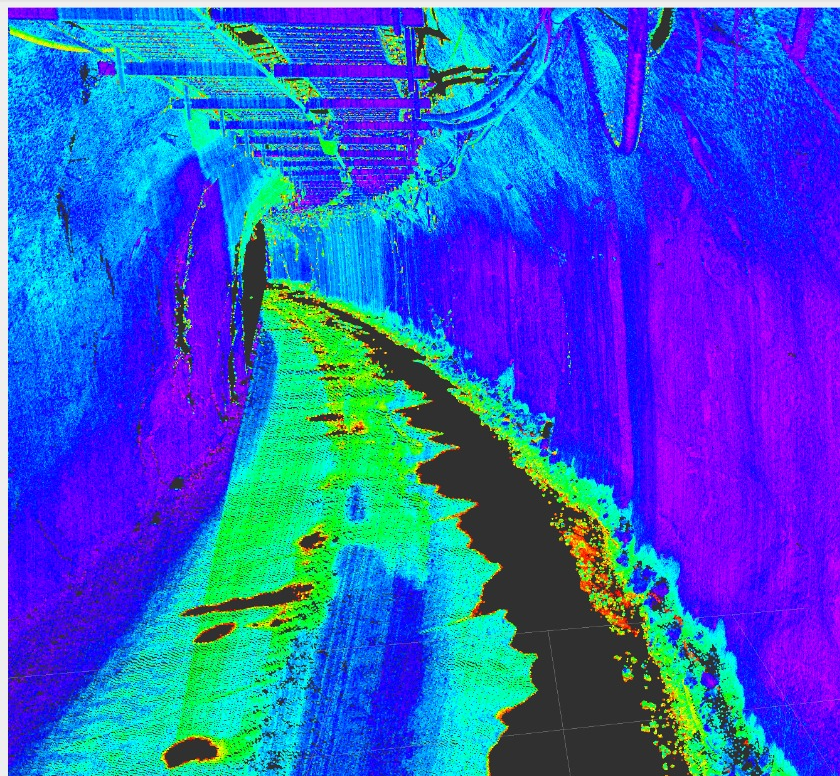}
  \caption{Point cloud visualisation from the Chilean underground mine
  survey. The irregular 3D tunnel geometry --- narrow corridors, branching
  junctions, and overhanging rock faces --- directly motivates the need
  for a physics-based benchmark grounded in real mine structure rather
  than procedural generation.}
  \label{fig:pointcloud}
\end{figure}

Fig.~\ref{fig:pointcloud} shows a representative point cloud from the
Chilean underground mine survey. We present MineXplore, a MuJoCo
navigation environment whose geometry is derived from the Leung et al.\
2017 dataset \cite{leung2017}. The environment exposes a
Gymnasium-style reset/step interface \cite{towers2024} and supports
training of reinforcement learning policies under local observations in a
topologically complex layout that procedural generators such as BARN do
not produce \cite{perille2020}. A full description of the
contour-to-MuJoCo XML Format (MJCF) compilation and validation pipeline
is provided in Section~\ref{sec:envdesign}.

\section{Related Work}

\subsection{Underground and Subterranean Robotics}

The DARPA Subterranean Challenge (2018--2021) drove recent progress in
autonomous systems for GPS-denied underground environments, with Team
CERBERUS winning the Final Event \cite{tranzatto2022,subt2021}. Ebadi et
al.\ catalogued open problems in extreme underground SLAM across six SubT
teams \cite{ebadi2024}. The Leung et al.\ 2017 Chilean underground mine
dataset, our geometric source, remains the only publicly released
production-mine sensor log suitable for navigation and SLAM research
\cite{leung2017}.

\subsection{Simulation Environments for Robot Learning}

Gazebo-based assets dominate the subterranean simulation niche: the SubT
Virtual Testbed ships replica competition courses \cite{subt2021}, and
BARN provides 300 procedurally generated navigation worlds used in ICRA
benchmarks \cite{perille2020}. For large-scale reinforcement learning,
Isaac Gym \cite{makoviychuk2021} and MuJoCo Playground \cite{zakka2025}
have become standard.

\subsection{Prior Use of the Chilean Mine Dataset}

Prior work has used the Leung et al.\ dataset exclusively for SLAM
benchmarking \cite{ebadi2024}. To the best of our knowledge, no prior
work has compiled its survey geometry into a physics-based simulation
environment.

\subsection{Real-Mine-Grounded Simulation}

Closest to our work, Gao and Awuah-Offei built a ROS~2 and Gazebo
framework for quadruped navigation in 3D maps of real sites including the
Edgar Mine teaching facility \cite{gao2025}. MuJoCo, and in particular
its GPU-vectorised MJX backend, supports parallel environment stepping at
rates orders of magnitude faster than Gazebo, consistent with the
throughput expectations of modern RL pipelines
\cite{makoviychuk2021,zakka2025}.

\subsection{Positioning of MineXplore}

\begin{itemize}
  \item \textbf{Real production-mine geometry.} Prior assets use
  procedural generation \cite{perille2020}, competition replicas
  \cite{subt2021}, or teaching facilities \cite{gao2025}. MineXplore is
  the first environment compiled from a real operational mine
  \cite{leung2017} inside MuJoCo/MJX.

  \item \textbf{Chilean dataset used for simulation.} All prior work uses
  Leung et al.\ \cite{leung2017} solely for SLAM benchmarking
  \cite{ebadi2024}; MineXplore is the first to convert it into a
  physics-based RL environment.

  \item \textbf{GPU-scalable over Gazebo.} Gazebo-based frameworks
  \cite{subt2021,gao2025} are impractical for deep RL training
  throughput; MuJoCo/MJX provides the parallel stepping rates required
  \cite{makoviychuk2021,zakka2025}.

  \item \textbf{Gymnasium interface over real mine topology.} Unlike BARN
  \cite{perille2020}, Isaac Gym \cite{makoviychuk2021}, or MuJoCo
  Playground \cite{zakka2025}, MineXplore exposes a standard reset/step
  interface over a 104{,}423\,m$^2$ loop-rich, real-mine tunnel network.
\end{itemize}

The Gymnasium-compatible interface design of MineXplore draws on
conventions established by representative simulation environments
including BARN \cite{perille2020}, Isaac Gym \cite{makoviychuk2021}, and
MuJoCo Playground \cite{zakka2025}, with the standard reset/step
interface provided by the Farama Gymnasium framework \cite{towers2024}.

\section{Environment Design}
\label{sec:envdesign}

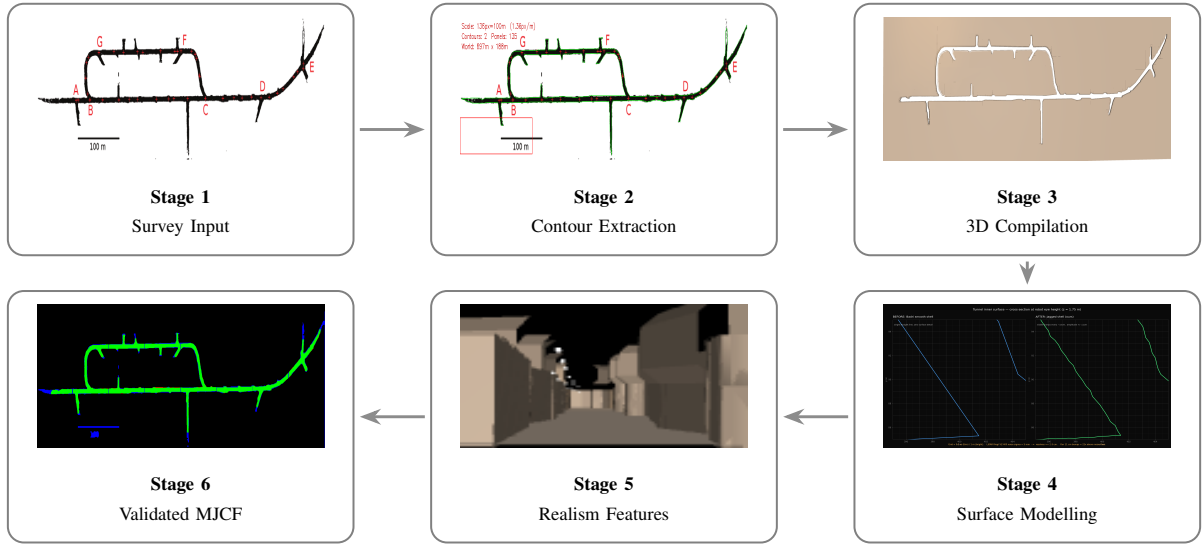
\begin{figure*}[t]
\centering
\begin{tikzpicture}[
  imgbox/.style={
    draw=black!50,
    rounded corners=6pt,
    fill=white,
    line width=0.75pt,
    inner sep=5pt
  },
  txtbox/.style={
    draw=black!50,
    rounded corners=6pt,
    fill=gray!6,
    line width=0.75pt,
    inner sep=8pt,
    minimum width=4.0cm,
    minimum height=3.8cm,
    align=center
  },
  myarrow/.style={
    ->,
    >=Stealth,
    line width=1.1pt,
    color=black!45,
    shorten >=2pt,
    shorten <=2pt
  }
]


\node[imgbox] (s1) at (0, 0) {%
  \begin{tabular}{c}%
    \includegraphics[width=3.8cm,height=\pipeImgHeight]%
      {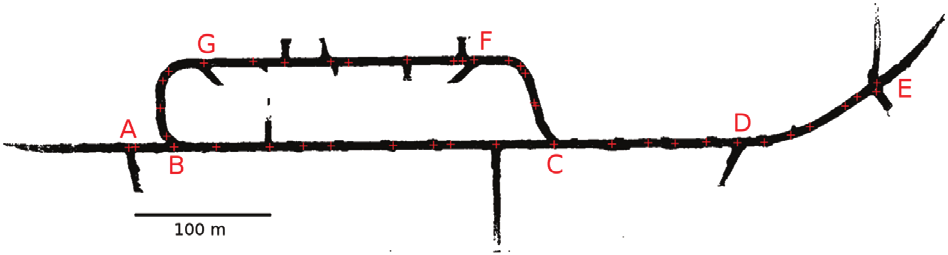}\\[4pt]%
    \textbf{\scriptsize Stage 1}\\[-1pt]%
    {\scriptsize Survey Input}%
  \end{tabular}%
};

\node[imgbox] (s2) at (5.6, 0) {%
  \begin{tabular}{c}%
    \includegraphics[width=3.8cm,height=\pipeImgHeight]%
      {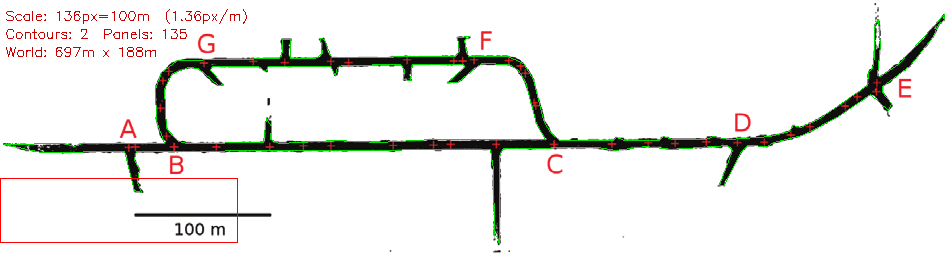}\\[4pt]%
    \textbf{\scriptsize Stage 2}\\[-1pt]%
    {\scriptsize Contour Extraction}%
  \end{tabular}%
};

\node[imgbox] (s3) at (11.2, 0) {%
  \begin{tabular}{c}%
    \includegraphics[width=3.8cm,height=\pipeImgHeight]%
      {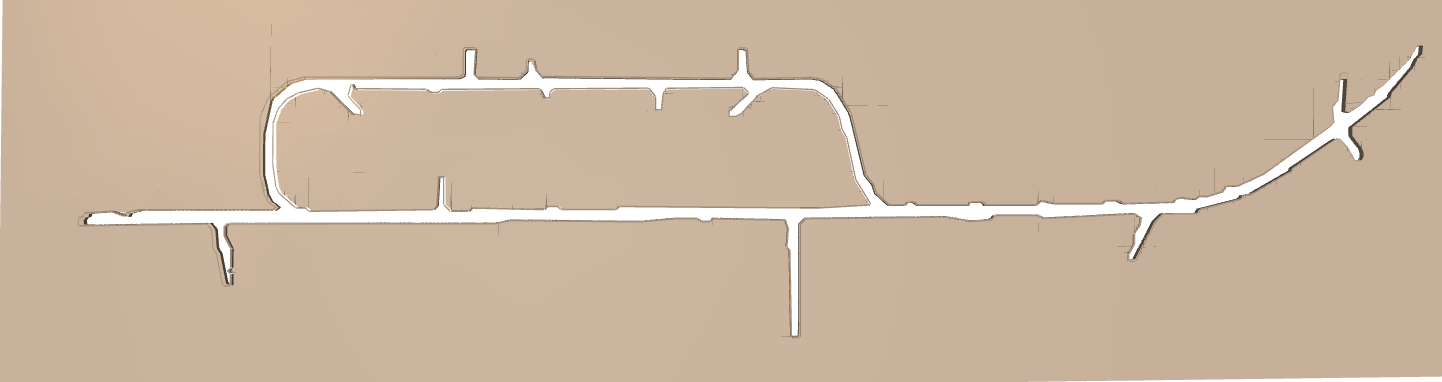}\\[4pt]%
    \textbf{\scriptsize Stage 3}\\[-1pt]%
    {\scriptsize 3D Compilation}%
  \end{tabular}%
};


\node[imgbox] (s4) at (11.2, -3.8) {%
  \begin{tabular}{c}%
    \includegraphics[width=3.8cm,height=\pipeImgHeight]%
      {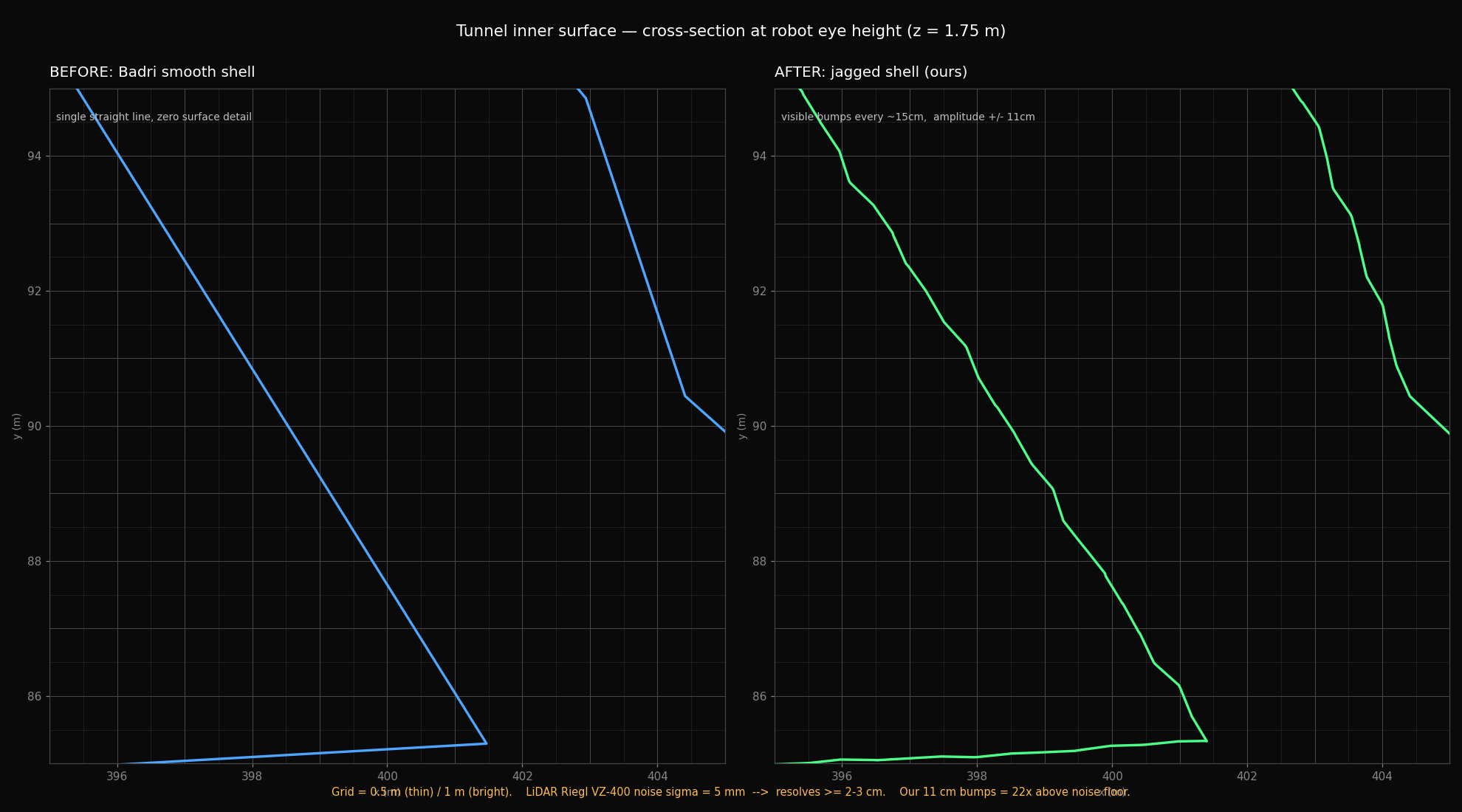}\\[4pt]%
    \textbf{\scriptsize Stage 4}\\[-1pt]%
    {\scriptsize Surface Modelling}%
  \end{tabular}%
};

\node[imgbox] (s5) at (5.6, -3.8) {
  \begin{tabular}{c}%
    \includegraphics[width=3.8cm,height=\pipeImgHeight]%
      {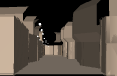}\\[4pt]%
    \textbf{\scriptsize Stage 5}\\[-1pt]%
    {\scriptsize Realism Features}%
  \end{tabular}%
};

\node[imgbox] (s6) at (0, -3.8) {%
  \begin{tabular}{c}%
    \includegraphics[width=3.8cm,height=\pipeImgHeight]%
      {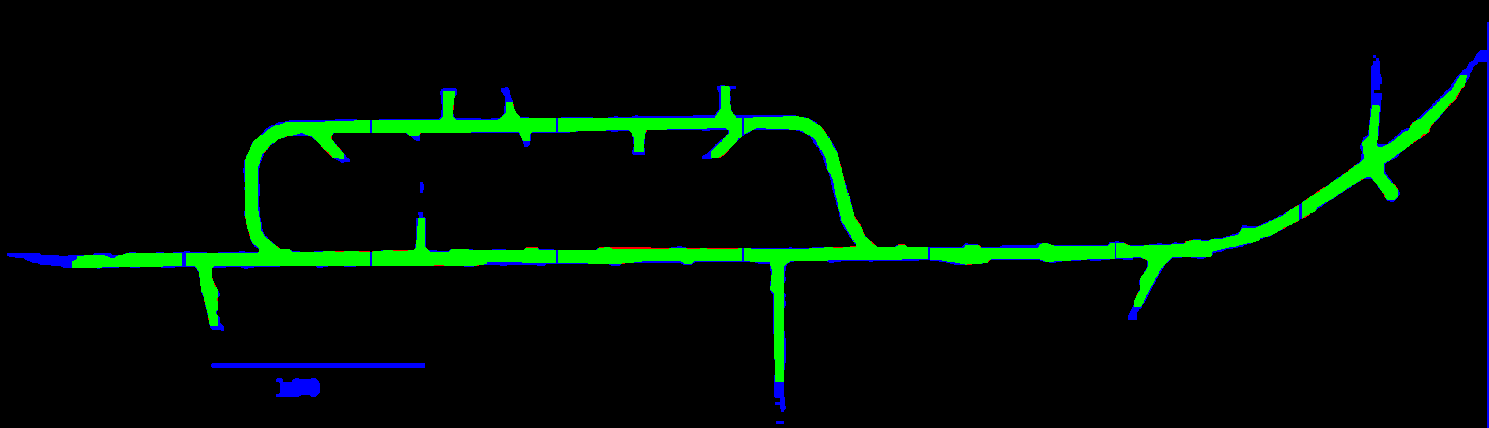}\\[4pt]%
    \textbf{\scriptsize Stage 6}\\[-1pt]%
    {\scriptsize Validated MJCF}%
  \end{tabular}%
};


\draw[myarrow] (s1.east) -- (s2.west);
\draw[myarrow] (s2.east) -- (s3.west);
\draw[myarrow] (s3.south) -- (s4.north);
\draw[myarrow] (s4.west) -- (s5.east);
\draw[myarrow] (s5.west) -- (s6.east);

\end{tikzpicture}
\caption{MineXplore compilation pipeline from real Chilean mine survey
data to a validated MJCF environment. Stage~1: 2D survey floor-plan and
LiDAR point cloud as source inputs. Stage~2: binarisation and OpenCV
contour detection of the tunnel boundary and rock island. Stage~3:
polygon triangulation, 3.5\,m extrusion, and V-HACD convex decomposition
into 1{,}186 collision geoms. Stage~4: two-layer visual wall system
comprising an 8-sided polygonal base layer and a LiDAR-sourced jagged
overlay. Stage~5: three realism features applied to the MJCF worldbody.
Stage~6: orthographic top-down IoU validation against the source survey
map (IoU\,$=$\,0.9538). Each stage produces a validated intermediate
output that feeds the next.}
\label{fig:pipeline}
\end{figure*}

Section~\ref{sec:envdesign} walks through the MineXplore compilation
pipeline in order: source inputs, 2D geometry extraction, 3D compilation,
surface feature and texture mapping, environment realism features, and
validation. Stage labels correspond to Fig.~\ref{fig:pipeline}.


\subsection{Source Data and Scale Calibration (Stage 1)}

\begin{figure}[H]
  \centering
  \includegraphics[width=\columnwidth]{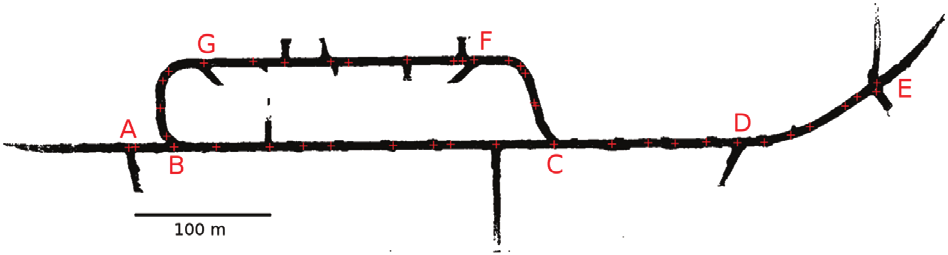}
  \caption{Published 2D survey floor-plan of the Chilean underground
  copper mine \cite{leung2017}. Junction density and corridor width in
  this source map directly determine navigation difficulty in the compiled
  environment --- tighter junctions and narrower passages increase the
  exploration challenge for the trained agent.}
  \label{fig:floorplan}
\end{figure}

MineXplore uses two complementary inputs \cite{leung2017}. The tunnel geometry is
derived from the published 2D survey floor-plan image shown in
Fig.~\ref{fig:floorplan}. The visual surface features---wall roughness,
cross-sectional profile, and surface colour---are derived from the
accompanying 3D LiDAR point cloud logs. We calibrated the
pixel-to-metric scale against the 100\,m scale bar on the source image,
yielding 1.36\,px/m. The nominal tunnel height is set to 3.5\,m,
consistent with the production tunnel specification in the source dataset.
The base floor geometry is a flat horizontal plane with no fine-grained
elevation mesh; cross-sectional height variation along the tunnel is not
modelled at the mesh level.

\subsection{2D Geometry Extraction (Stage 2)}

\begin{figure}[H]
  \centering
  \includegraphics[width=\columnwidth]{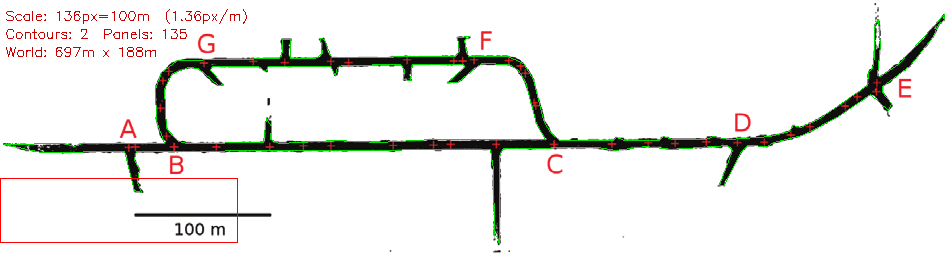}
  \caption{Detected contours overlaid on the binarised survey map. Outer
  tunnel boundary in green; interior rock island in red. Accurate contour
  capture at this stage controls wall placement in the compiled model ---
  the green outer boundary and red rock island correctly enclose the
  navigable tunnel area, with no false closures or missed passages.}
  \label{fig:contours}
\end{figure}

We binarised the survey floor-plan and extracted the tunnel footprint
using OpenCV contour detection. The largest external contour defines the
outer tunnel boundary; interior contours define rock islands and pillars.
Fig.~\ref{fig:contours} shows the detected contours overlaid on the
binarised map. Each contour was converted to a Shapely polygon with holes
and simplified using Douglas--Peucker with a tolerance tuned to preserve
corners under 1\,m. The natural irregularity retained by this tolerance
is the primary source of jagged wall geometry in the compiled model. The
compiled footprint contains one outer boundary and one dominant interior
rock island of 14{,}577\,m$^2$, yielding a navigable area of
104{,}423\,m$^2$ within a world extent of approximately 697\,m $\times$
188\,m.

\subsection{3D Compilation and Physics Setup (Stage 3)}

\begin{figure}[H]
  \centering
  \includegraphics[width=\columnwidth]{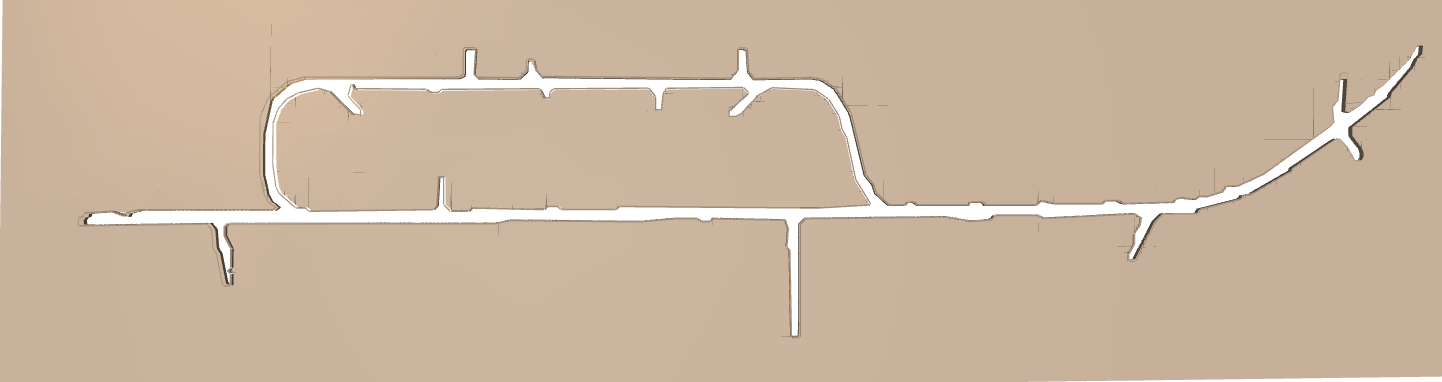}
  \caption{Compiled MineXplore environment in the MuJoCo interactive
  viewer. Confirms that the geometry
  pipeline introduces no systematic distortions to tunnel structure or
  junction connectivity.}
  \label{fig:viewer}
\end{figure}

We triangulated the 2D polygon set and extruded each face to 3.5\,m. MuJoCo's collision engine requires convex geoms; we decomposed
the extruded mesh using V-HACD with a resolution of 1{,}000{,}000, a
concavity threshold of 0.5, and a maximum of 32 convex hulls per rock
tile. Applied across 70 rock geometry tiles, this produces 1{,}186 convex
collision geoms in the final MJCF model. The compiled world is loaded and
centred on the main tunnel network. Ceiling and floor planes
close the tunnel volume. Physics parameters set baseline wall
and floor friction and include an occupancy octomap for collision-free
spawn sampling. Compilation and scene load complete in under ten seconds
on an AMD Ryzen 5 5600H under Ubuntu Linux. Fig.~\ref{fig:viewer} shows
the compiled MuJoCo world.

\subsection{Surface Wall Geometry and Visual Detail (Stage 4)}
\label{sec:surface}

\begin{figure}[H]
  \centering
  \includegraphics[width=\columnwidth]{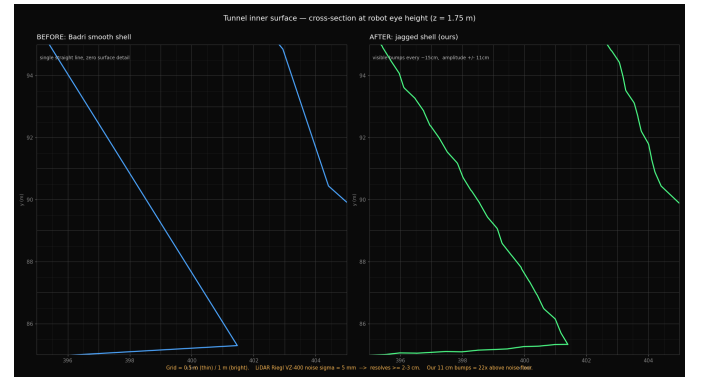}
  \caption{Cross-sectional wall profile comparison: smooth flat-extruded
  geometry (left) versus the two-layer surface model (right). The
  two-layer model adds wall irregularity absent in the flat-extruded
  baseline; higher wall surface complexity increases LiDAR return
  variation, making sensor-based navigation harder and more
  representative of real mine conditions.}
  \label{fig:crosssection}
\end{figure}

The base layer is an 8-sided polygonal mesh that approximates the
characteristic bore cross-section of the production tunnel. The overlay
layer is a jagged mesh derived from the Leung et al.\ 3D LiDAR point
cloud that captures the centimetre-scale rock surface irregularity of the
real tunnel walls. Both mesh geoms are configured as purely visual and do
not participate in collision detection; physics contact is handled
exclusively by the V-HACD convex hull decomposition.
This two-layer design separates physical accuracy from visual realism and
avoids the computational cost of simulating fine-surface contact. A
uniform rock material is applied across all 1{,}186 rock meshes and both
wall layers. Fig.~\ref{fig:crosssection} illustrates the difference
between the original flat-extruded geometry and the two-layer surface
model.

\subsection{Environment Realism Features (Stage 5)}

MineXplore incorporates three additional features that replicate the
physical conditions of the Chilean copper mine.

\textit{Terrain incline.} The entire MJCF worldbody is rotated
5$^\circ$ (0.0872665\,rad) around the $Y$-axis, tilting walls, floor,
rocks, and lights uniformly as a rigid body. This produces an effective
elevation gain of 5.2\,m across the tunnel's entrance region, consistent
with the terrain gradient in the source dataset, without modifying the
V-HACD collision decomposition.

\textit{Terrain friction zones.} Three spatially fixed floor box geoms
partition the tunnel along its east--west axis to replicate mine water
seepage: a dry western zone (friction 1.00), a muddy central zone
(friction 0.55), and a waterlogged eastern zone (friction 0.25), each
covering 119\,m $\times$ 102\,m. The contact friction compounds with
the actuation-level scaling in Section~\ref{sec:benchmark} to jointly
model traction degradation across ground conditions.

\textit{Periodic lighting.} Thirty-three MuJoCo spot lights are placed
along the tunnel centreline at 20\,m intervals, 3.0\,m above the floor,
directed slightly downward with a 45$^\circ$ cone cutoff.

\subsection{Geometry \& Surface Validation  (Stage 6)}
\begin{figure}[H]
  \centering
  \includegraphics[width=\columnwidth]{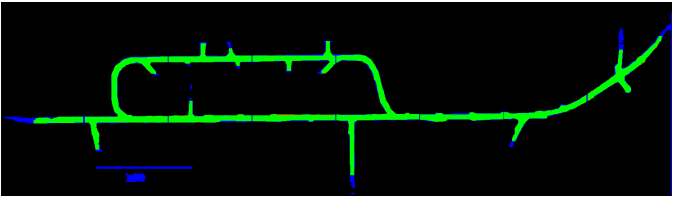}
  \caption{Pixel-wise geometric fidelity overlay. Green pixels indicate
  agreement; blue pixels indicate deviation. Green pixels (95.3\%)
  confirm geometric agreement with the source map; blue deviation pixels
  are concentrated at junctions and corridor edges as expected from the
  Douglas--Peucker simplification, and are not distributed randomly across
  the tunnel interior.}
  \label{fig:iou}
\end{figure}

We validated the compiled model on two axes. For navigability, we cast
rays through the free volume and confirmed that every pixel marked
navigable in the source floor-plan maps to an unobstructed column in the
MJCF model. For geometric consistency, we rendered an orthographic
top-down projection of the compiled MJCF model, binarised it, and
compared it pixel-wise against the original survey floor-plan.
Fig.~\ref{fig:iou} shows the resulting overlay: green pixels mark
agreement; blue pixels mark deviation, concentrated at junctions and
corridor edges where the Douglas--Peucker simplification rounds tight
corners. The strict Intersection over Union (IoU) after noise cleanup is
0.9538, confirming that the compiled model reproduces 95.3\% of the
source survey footprint at pixel level. This overlay is produced directly
from the compiled MJCF, not from any intermediate representation.

\begin{figure}[H]
  \centering
  \includegraphics[width=\columnwidth]{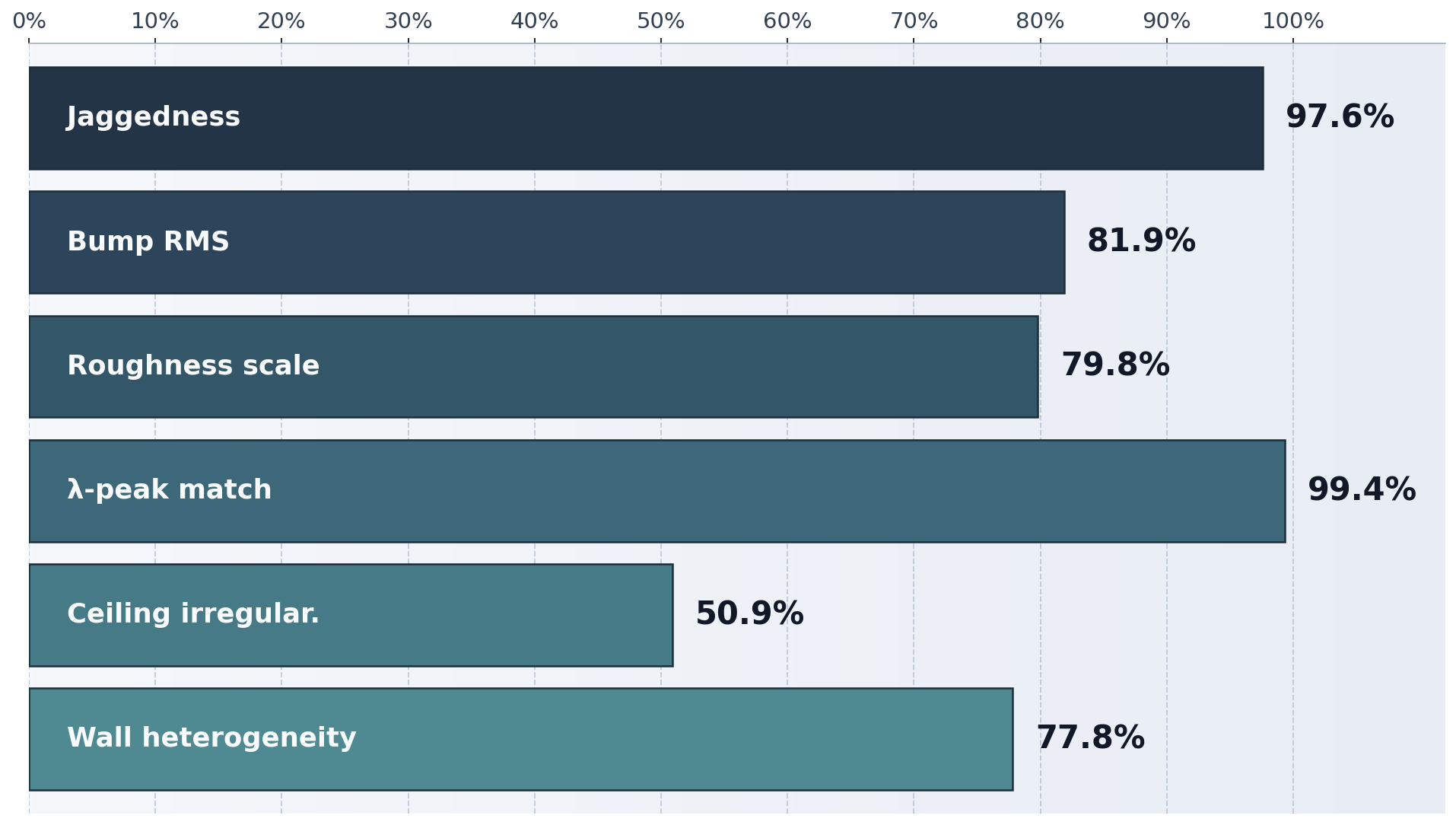}
  \caption{Per-axis surface-texture similarity comparison between the
  compiled MineXplore model and the real Chilean mine LiDAR point cloud.
  All six dimensions score above 50\%; dominant wavelength (99.4\%) and
  wall jaggedness (97.6\%) are near-perfect matches, confirming that the
  spatial frequency structure of the real mine walls is faithfully
  reproduced at the scales most relevant for LiDAR-based navigation.}
  \label{fig:texture}
\end{figure}

We assessed surface texture fidelity by comparing the compiled MineXplore
wall geometry against the real 3D LiDAR point cloud by fitting SVD-based
principal planes over a voxel-scale sweep $w \in \{0.10, \ldots,
2.00\}$\,m, computing ISO~4287-style residuals ($R_a$, $R_q$, $R_z$,
$p_{95}$), and extracting a 1D along-wall wavelength spectrum via
arc-length-resampled FFT across 5\,cm horizontal slabs. Six structural
dimensions were scored on a per-axis similarity metric
(Fig.~\ref{fig:texture}): wall jaggedness (97.6\%), bump $R_q$ (81.9\%),
roughness scale integral (79.8\%), dominant wavelength $\lambda$-peak
(99.4\%), ceiling irregularity (50.9\%), and wall heterogeneity (77.8\%),
yielding an overall geometric-mean score of 79.4\%. The simulated surface
undershoots real roughness magnitude by ${\approx}2.5\times$ at the
half-metre scale---an intentional amplitude clip to prevent narrow
passages from pinching shut---while faithfully replicating the dominant
5--20\,cm spatial-frequency band of the real mine walls.

\section{Benchmark Interface}
\label{sec:benchmark}

\subsection{Observation Space}

Each agent receives a 19-dimensional observation vector at every control
step. The first 16 elements are LiDAR range readings from a simulated 2D
scanner with 16 uniformly spaced beams covering the full 360$^\circ$
field of view ($[0, 2\pi)$). Raw ranges are clipped to
$[0.12\,\text{m},\, 10.0\,\text{m}]$, then normalised to $[0, 1]$.
Gaussian noise with $\sigma = 0.01$\,m is applied to each beam before
normalisation. The remaining 3 elements are: the previous normalised
linear command, a fixed scalar of 0.0, and the previous normalised
angular command, providing the agent with one step of action history.

\subsection{Action Space}

The action space is a continuous 2D box $\mathcal{A} = [-1, 1]^2$. The
policy output $(a_0, a_1)$ is mapped to differential-drive commands as:
\begin{equation}
  v = a_0 \cdot 1.5\,\text{m/s}, \qquad \omega = a_1 \cdot
  1.5\,\text{rad/s}
\end{equation}
Commands are then scaled and perturbed by terrain type before execution:
dry terrain applies scale 1.0 with zero noise; muddy terrain applies
scale 0.7 with additive noise $\sigma = 0.04$\,m/s; waterlogged terrain
applies scale 0.4 with noise $\sigma = 0.10$\,m/s. This actuation-level
scaling compounds with the MuJoCo contact friction values described to model the full traction degradation across terrain types.

\subsection{Episode Structure}

MineXplore is a pure exploration environment. There is no goal position:
\texttt{terminated} is always \texttt{False} and episodes end only by
truncation. The maximum episode horizon is 6{,}000 control steps at a
timestep of 0.1\,s, giving a nominal maximum duration of 600\,s. An
episode is also truncated early if the agent's speed remains below
0.01\,m/s for 100 consecutive steps, preventing degenerate stationary
policies. At the start of each episode, the agent is spawned at a fixed
anchor near the mine entrance; if that cell fails clearance checks, the
agent falls back to a random valid point in the entrance region with
uniform yaw sampling over $[-\pi, \pi)$.

\subsection{Reward Structure}
\label{sec:reward}

The per-step reward $r$ is:
\begin{equation}
  r = \Delta c \cdot g
    - \mathbbm{1}[\Delta c = 0] \cdot 0.002
    + M
    - \mathbbm{1}[\text{collision}] \cdot 15.0
\end{equation}

\noindent The reward is structured around a dense exploration signal
combined with hard safety and efficiency constraints. Without a per-step
coverage term, initial experiments showed coverage saturating near 40\%:
the agent circled already-mapped corridors since no per-step incentive
existed to push into undiscovered regions, producing a flat reward curve
beyond early training. The $\Delta c \cdot g$ term resolves this by
rewarding only \textit{newly} discovered cells at each step --- using a
delta rather than cumulative coverage ensures the agent receives signal
exclusively for active exploration and not for revisiting mapped space.
The per-cell gain $g = 10.0 / C_{\text{total}}$ normalises reward
magnitude by total tunnel area, making it resolution- and
map-size-invariant. The step penalty $-\mathbbm{1}[\Delta c=0]\cdot
0.002$ discourages stationary behaviour. Milestone bonuses $M$ at
25\%~($+15$), 50\%~($+30$), and 75\%~($+45$) provide curriculum-style
waypoints that prevent the policy stalling at local coverage maxima
during early training. The collision penalty of $-15.0$ is set at
$7{,}500\times$ the step penalty, creating an unambiguous priority
ordering: wall avoidance dominates over time efficiency, which dominates
over standing still. We use RL solely to validate that MineXplore
supports stable exploration under local observations; it is not a
contribution of this paper.

\subsection{Reproducibility}

In the experiments reported here we use a single-agent configuration.
All experiments use MuJoCo 3.2.6 \cite{todorov2012}. Training is run
across five independent random seeds (42, 142, 242, 342, 442); all
metrics are reported as mean $\pm$ standard deviation across seeds. The
compiled MJCF model is kept under version control. Code and environment
pipeline will be made open-source upon publication.

\section{Results}

\subsection{Reinforcement Learning Baseline}

We use a single-agent PPO policy, trained via RLlib, to validate that MineXplore supports stable
exploration learning under local observations. RL is not a contribution of
this paper; the baseline exists solely to confirm that the environment is
navigable and that reward signals are well-formed. Training was run across
five independent random seeds with the reward structure defined in
Section~\ref{sec:reward}. Per-seed episode counts varied by stopping
condition: three seeds (42, 342, 442) terminated upon reaching the coverage
target at episodes 257, 432, and 282 respectively; two seeds (142, 242)
triggered the early-stop criterion after coverage regression at episodes
962 and 416. All metrics are reported as mean $\pm$ standard deviation
across all five seeds over the final 100 episodes of each run.

\begin{table}[H]
\centering
\caption{Per-seed training outcomes. Stop: TC\,=\,target coverage
reached; ED\,=\,early stop due to coverage drop.
BestRoll\,=\,best rolling coverage.}
\label{tab:seeds}
\begin{tabular}{ccccc}
\toprule
Seed & Stop & Ep. & Final Cov.\ (\%) & BestRoll (\%) \\
\midrule
42  & TC & 257  & 100.00 & 89.79 \\
142 & ED & 962  & 70.05  & 85.43 \\
242 & ED & 416  & 70.77  & 89.39 \\
342 & TC & 432  & 89.83  & 89.95 \\
442 & TC & 282  & 92.00  & 89.88 \\
\midrule
Mean & -- & -- & $84.53 \pm 12.02$ & $88.89 \pm 1.74$ \\
\bottomrule
\end{tabular}
\end{table}

Table~\ref{tab:seeds} summarises the per-seed outcomes. Three seeds
(42, 342, 442) reached the coverage target and stopped cleanly. Two seeds
(142, 242) triggered early-stop due to coverage regression. Despite early
stopping, all five seeds achieved a best rolling coverage between 85.43\%
and 89.95\%, giving an aggregate best rolling coverage of
$88.89\% \pm 1.74\%$.

\begin{figure}[H]
  \centering
  \includegraphics[width=\columnwidth]{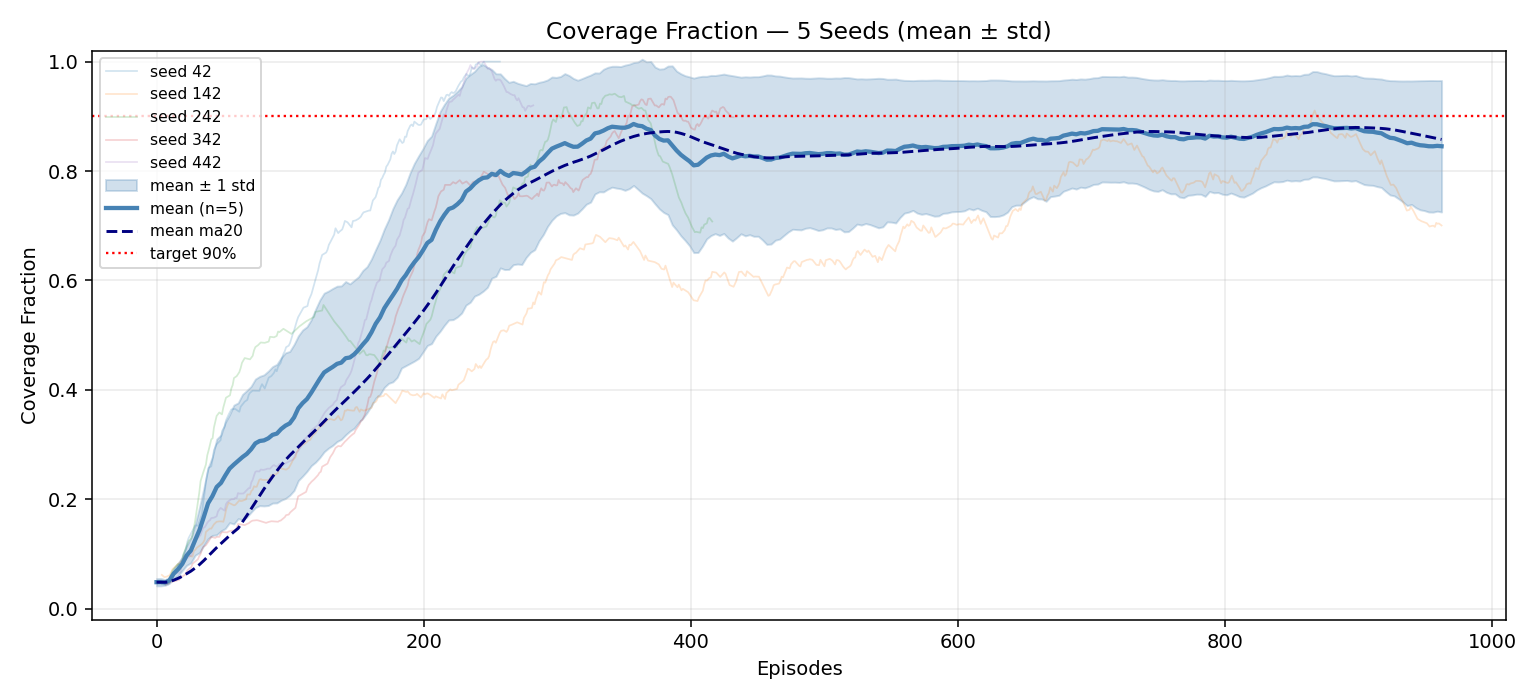}
  \caption{Coverage fraction over training (mean $\pm$ 1\,SD across
  five seeds; dotted line: 90\% target). Mean rises monotonically
  from ${\sim}5\%$ to a best rolling average of $88.89\% \pm 1.74\%$;
  higher is better.}
  \label{fig:coverage}
\end{figure}

Fig.~\ref{fig:coverage} shows the mean coverage fraction across all five
seeds. The mean rises monotonically from ${\sim}5\%$ at episode~1 and
approaches the 90\% target threshold, with the shaded band confirming
consistent behaviour across seeds. The two early-stopped seeds account
for the slight downward drift in the mean after episode~400, which is
itself informative: even without reaching the target, those seeds
sustain above-85\% best rolling coverage.

\begin{figure}[H]
  \centering
  \includegraphics[width=\columnwidth]{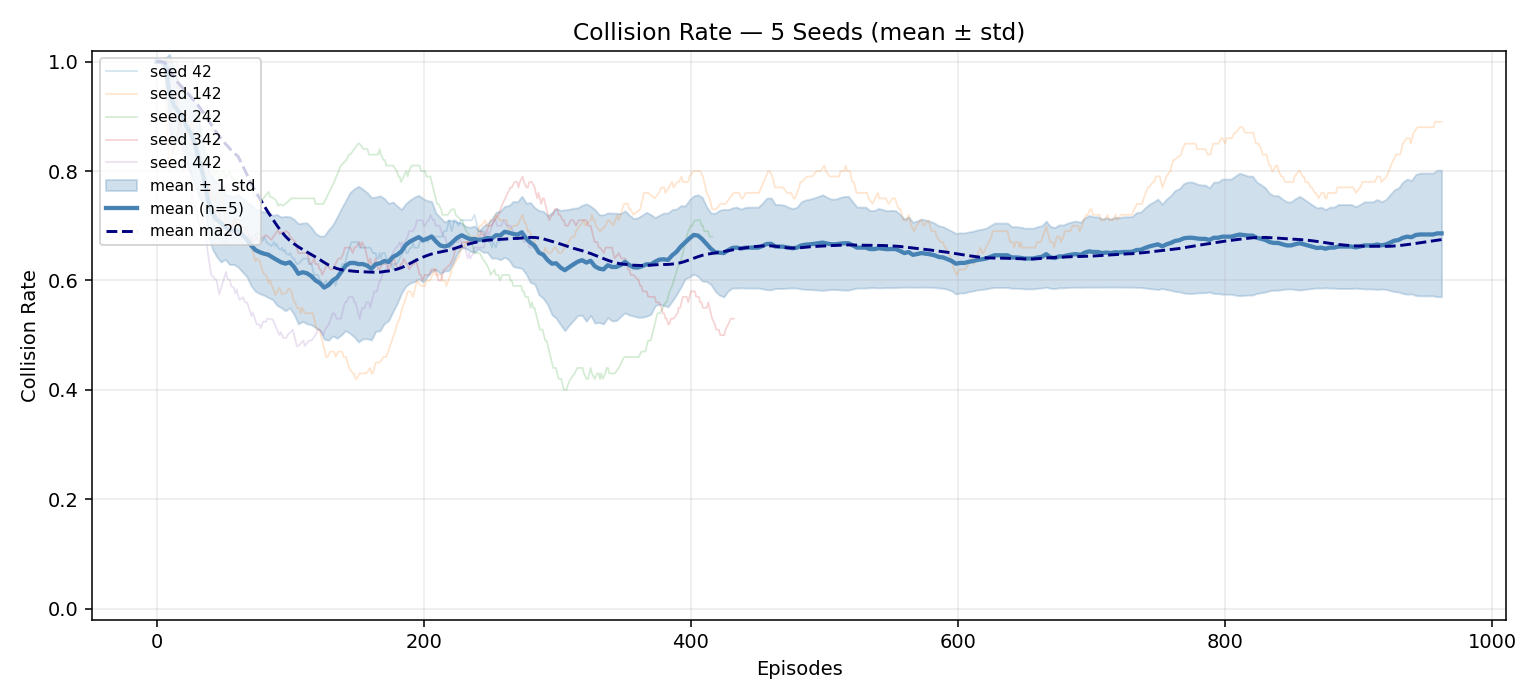}
  \caption{Per-episode collision rate over training (mean $\pm$ 1\,SD
  across five seeds). Mean falls from ${\sim}1.0$ to $0.65 \pm 0.13$
  in the final episodes; lower is better.}
  \label{fig:collision}
\end{figure}

Fig.~\ref{fig:collision} shows the per-episode collision rate. The mean
rate begins near 1.0 and falls to $0.65 \pm 0.13$ in the final episodes.
The persistent collision rate reflects the difficulty of navigating
V-HACD collision geometry under purely local LiDAR observations; the
declining trend across all five seeds confirms that the environment
produces a consistent learning signal for wall avoidance.

\begin{figure}[H]
  \centering
  \includegraphics[width=\columnwidth]{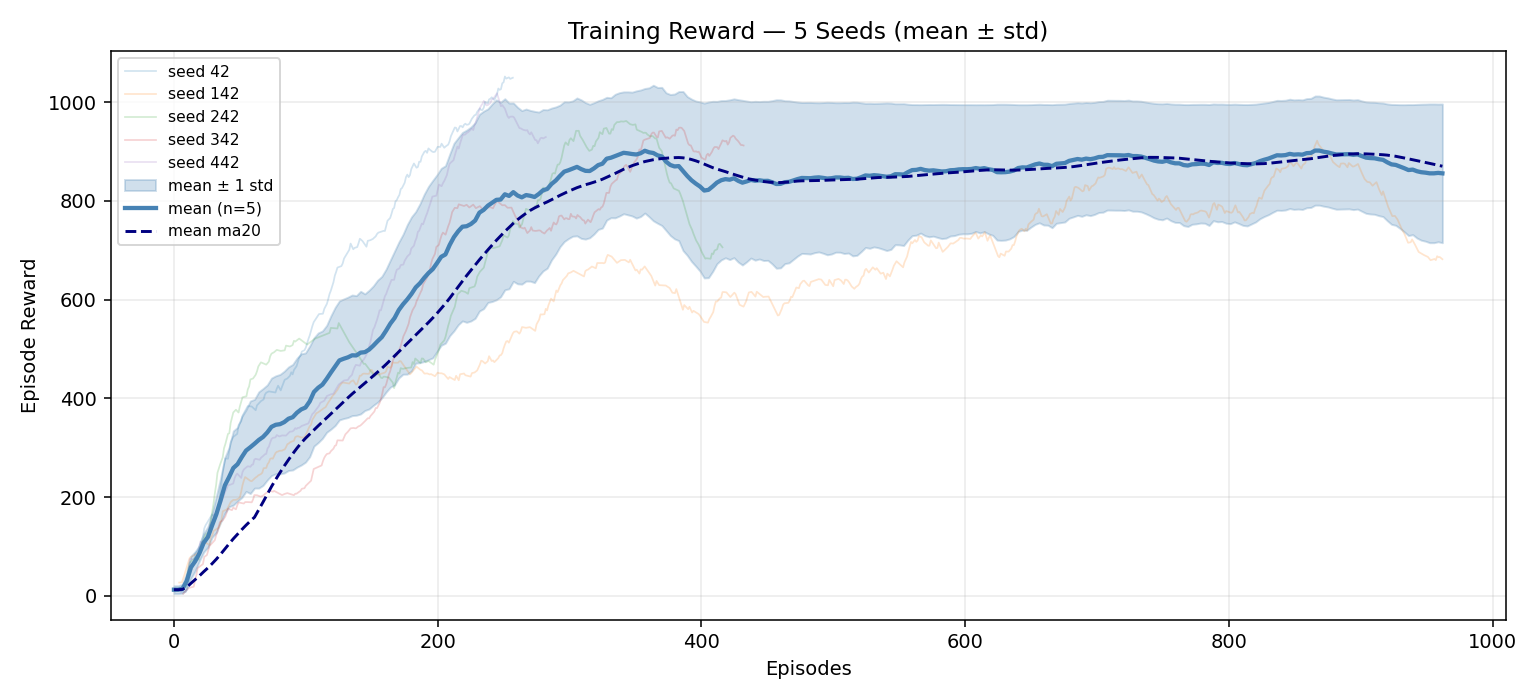}
  \caption{Cumulative episode reward over training (mean $\pm$ 1\,SD
  across five seeds). Rises monotonically to $860 \pm 150$; higher is
  better; alignment with coverage confirms no degenerate exploitation
  across any seed.}
  \label{fig:reward}
\end{figure}

Fig.~\ref{fig:reward} shows episode reward, which rises monotonically
and stabilises around $860 \pm 150$. The alignment with coverage growth
confirms that reward growth is coupled to exploration breadth as designed,
with no degenerate reward exploitation visible across any of the five seeds.
\section{Discussion}

MineXplore is a single-mine MuJoCo/MJX navigation environment grounded
in a real production tunnel, targeted at researchers who need a
non-procedural, topology-rich underground test case for navigation and
exploration algorithm development. It is not a physics-calibrated digital
twin: wall friction, surface roughness, and contact stiffness are
approximations derived from the source data, not direct measurements. The
collision model is a V-HACD convex decomposition, which may introduce localised
hull artefacts in the narrowest tunnel passages. The elevation model is a
single global rotation approximating the tunnel gradient; fine-grained
topographic variation and cross-sectional height changes along the full
tunnel length are not modelled and are deferred to future work using the
Leung et al.\ 3D LiDAR elevation logs. The five-seed evaluation reveals
meaningful variance in coverage outcomes (${\pm}12.02\%$ final,
${\pm}1.74\%$ best rolling), with two seeds triggering early stops due to
coverage regression --- a property that distinguishes strong from weak
exploration initialisations and is desirable in a benchmark intended to
stress-test exploration algorithms.

MineXplore is the right tool when a large-scale, real-mine-derived
obstacle topology inside a MuJoCo or MJX training pipeline is needed,
alongside a Gymnasium-compatible interface for single- or multi-agent
navigation under local observations, or matched source data for future
sim-to-real work grounded in the Leung et al.\ Chilean mine dataset
\cite{leung2017}. Future work includes full elevation-aware geometry from
the Leung et al.\ 3D LiDAR logs, LiDAR-replay validation against the
original dataset sweeps, procedural cross-section perturbation for
environment diversity, broader multi-agent and cross-algorithm
baselines \& point cloud data verification.

\section{Conclusion}

We presented MineXplore, an open-source MuJoCo navigation environment
derived from the Leung et al.\ 2017 Chilean underground copper mine
survey. We described the six-stage contour-to-MJCF compilation pipeline
(Fig.~\ref{fig:pipeline}), validated the compiled geometry against the
source survey via a top-down overlay (Fig.~\ref{fig:iou}), and confirmed
navigability using a single-agent PPO baseline that
achieves a best rolling coverage of $88.89\%\ {\pm}\ 1.74\%$ with a mean
final coverage of $84.53\%\ {\pm}\ 12.02\%$ across five independent
seeds. MineXplore closes a specific gap in the robot-learning tooling
landscape: a real-mine-grounded, Gymnasium-compatible navigation
environment in the MuJoCo ecosystem, augmented with LiDAR-sourced wall
geometry, terrain heterogeneity, a global incline, and periodic spot
lighting. Future work includes full elevation-aware geometry from the
Leung et al.\ 3D LiDAR logs, LiDAR-replay validation against the
original dataset sweeps, procedural cross-section perturbation for
environment diversity, and broader multi-agent and cross-algorithm
baselines.

\section*{Acknowledgment}
The authors thank BuildMachineLabs for providing the opportunity,
research direction, and resources that supported this work. The authors
also thank Leung~et~al.\ for releasing the Chilean underground mine
dataset~\cite{leung2017}, which forms the geometric basis of MineXplore.

\bibliographystyle{IEEEtran}

\end{document}